\documentclass[11pt,letterpaper]{article}
\usepackage{cogsys}
\usepackage{graphicx}
\usepackage[T1]{fontenc}
\usepackage{times}
\usepackage{xcolor}

\usepackage{mathtools}
\usepackage{enumitem}
\usepackage{amsmath}
\usepackage{amssymb}
\usepackage{subcaption}
\usepackage{natbib}
\newcommand{\noskip}{\vspace{-\parskip}}	

\ShortHeadings{Plant Inspired Problems}
              {D.\ Sanyal, J.\ Michelson, C.\ Cao, A.\ Roddy, and M.\ Kunda}

\begin{document} 

\title{Plant-Inspired AI: Plants as Inspiration for Novel Problem Formulations, and Two Case Studies}
 
\author{Deepayan Sanyal\footnote[1]{Both contributed equally}}{deepayan.sanyal@nyu.edu}
\author{Joel Michelson\textsuperscript{1}}{j.michelson@nyu.edu}
\address{Department of Environmental Studies, New York University, New York City, NY, 10003, USA}
\author{Carla E. Cao}{carla4ecao@gmail.com}
\address{Department of Philosophy, Universidad de Murcia, Murcia, 30010, Spain}
\author{Adam B. Roddy}{adam.roddy@nyu.edu}
\address{Department of Environmental Studies, New York University, New York City, NY, 10003, USA}
\author{Maithilee Kunda}{mkunda@ed.ac.uk}
\address{School of Informatics, University of Edinburgh, Edinburgh, Scotland, EH8 9AB, UK}
\vskip 0.05in
 
\begin{abstract}
Artificial Intelligence (AI) has long been inspired by studies of biological intelligence. Reinforcement learning, for instance, drew inspiration from studies involving animal learning and is now a powerful paradigm for solving many real-world problems. Recently, plant biologists have uncovered a wide range of complex behaviors in plants that enable them to flexibly adapt to variable environments. Here, we argue that such behavior can motivate new AI frameworks encompassing a range of problems overlooked by existing problem-solving frameworks such as supervised learning, tree search, and constraint satisfaction. We illustrate this idea with two examples of intelligent problem-solving in plants: (1) leaf mimicry in \textit{Boquila trifoliolata}, a vine capable of altering its leaves' morphology to resemble those of multiple host trees simultaneously; and (2) coordinated root-shoot growth, wherein plants allocate resources across organ systems exploring distinct environments. While leaf mimicry is highly specific to \textit{Boquila}, coordination of root-shoot growth is shared across most plants. For both examples, we capture underlying computational principles and identify problems fitting these frameworks that are currently unaddressed by AI. Finally, we outline preliminary task formulations and discuss how these formulations may be applied to non-plant problems.
\end{abstract}

\section{Introduction} 

AI research has provided many tools for society that are used in a wide variety of applications. For example, reinforcement learning (RL) provides a general framework for sequential decision-making and has been applied to a wide range of problems from robotic control and chatbot development to financial decision-making and drug discovery. What makes RL so powerful? Over the decades, AI researchers have developed exceedingly clever algorithms for RL problems, such as TD-Learning, SARSA, PPO, and more. In addition to these RL algorithms, part of what makes RL so powerful is its \textbf{problem formulation} (PF). In AI, a PF is a structured collection of mathematical variables that defines an abstract problem.  For a given PF, we can attempt to create a \textit{solution algorithm} that solves it, and a given algorithm can be run on concrete \textit{instances} of the PF to produce concrete \textit{solutions}. A major reason we can apply an algorithm like PPO across so many different applications (robotics, drug discovery, etc.) is because the underlying RL PF itself can be fit onto so many different real-world instances. RL is not unique in this respect: AI has yielded many other powerful PFs like constraint satisfaction, shortest path problems, and supervised learning, each of which provides a common abstraction over a broad set of problems.

This raises a fundamental question: \textit{Where do these powerful PFs like RL or constraint satisfaction come from?} They do not exist \textit{a priori} in nature or in mathematics.  Rather, PFs are mathematical artifacts that must be invented by AI researchers: abstractions that capture useful structure in a class of problems. 
One source of PFs is an abstract model of a concrete real-world problem that needs solving. For instance, many PFs in control theory were originally developed to address physical engineering problems \citep{bennett1993development}. Over time, these PFs can become sufficiently general that their use extends beyond the problems that originally motivated them. Thus, identifying new abstractions that capture useful structure applicable to a wide range of problems is a crucial part of AI research. \textit{Where should we look for such abstractions?}

Interestingly, some of AI's most powerful and widely used PFs were inspired by problems solved by biological intelligence, i.e., by humans or animals. For example, the RL PF was inspired in part by studies of trial-and-error learning in animals \citep{sutton1998reinforcement}.  The shortest path PF, another very old and widely used PF---and today used for much, much more than finding literal paths---was first formally defined in the late 1800s based on the idea of a person navigating a maze \citep{schrijver2012history}. Outside of AI, research on design (e.g., engineering or industrial design) includes highly relevant work on how problem formulations originate and co-evolve alongside solutions \citep[e.g.,][]{volkema1983problem,maher1996modeling,crilly2021evolution}, though in these contexts, PFs usually refer to something more specific than they do in AI, like the requirements for a specific product. 

In this paper, we 
argue that a rich source of inspiration for new AI PFs lies in the problems solved by the diversity of plant life and that these phenomena can be studied through the lens of computation. Plants are particularly interesting sources of PFs because of how their capabilities and constraints differ fundamentally from animals. Plants lack animals' centralized nervous systems and hence need to coordinate resource acquisition and growth through distributed chemical processes. Being sessile, their main mode of responding to and adapting to environmental changes is through modulating their biochemistry, physiology, growth, and development. Plants are also some of the most evolutionarily successful organisms on Earth, making up the majority of global biomass \citep{bar2018biomass} and employing a diverse set of adaptations for survival \citep{singhal2021diversification}.

Considering plants as intelligent and expressing cognitive capacities has been fraught with debate \citep{gagliano2018plants, taiz2019plants, calvo2020plants, mallatt2023plant}. Regardless of whether the scientific community attributes `intelligence' to plant phenomena, a biological behavior need not be intelligent for it to be a useful object of study for AI. Interesting problems arise whenever a system with finite resources must allocate those resources to accomplish a goal under constraints.

For these reasons, we expect that studying the problems that plants solve will inspire new AI PFs that consider fundamentally different means of perception, communication, and decision-making. In this paper, we explore this potential through two intriguing examples of plant decision-making that suggest novel PFs. 
Specifically, we:

\begin{itemize}
    \noskip \item Present a novel PF inspired by leaf mimicry in \textit{Boquila trifoliolata}, and list its novel key problem characteristics including substrate plasticity and relative localization, in Section \ref{ref:pf1}; 
    \noskip \item Present a novel PF inspired by the cooperation between a large number of root tips and a large number of shoots common to most plants and list the key problem characteristics of this PF including resource allocation and heterogeneous roles, in Section \ref{ref:pf2};
    \noskip \item Describe specific computational tasks in these two case studies that remain unaddressed by AI PFs (Sections \ref{ref:mimic} and \ref{ref:rootsnshoots});
    \noskip \item Discuss limitations in our methods of PF and task generation, how future work could address the PFs shown here, in Section \ref{ref:discussion}, and PFs inspired by different plant phenomena, in the Appendix.
\end{itemize}

\section{Related Work: Computational Modeling of Plant Behavior}

Challenging traditional views that plants are unintelligent, 
\citet{trewavas2003aspects} argued that plants perceive aspects of their environment and respond flexibly to them. They discuss plant signal transduction, environmental discrimination, phenotypic plasticity, learning and memory, while also raising questions about whole plant behavior. Later work has developed these new ideas into broader accounts of plant behavior, including conceptual frameworks for understanding plants as information-processing systems \citep{segundo2025plant, segundo2026plant}. 

Several studies have proposed computational mechanisms that could underlie complex plant behavior. \citet{calvo2016feature} consider the value of describing plant behavior with either feature detection-based or predictive coding-based models and proposed empirical approaches for comparing such models. \citet{calvo2017predicting} further develop this idea using the free-energy principle and active inference, arguing that plants proactively sample their environment and use predictions about sensory stimulation to guide adaptive behavior. \citet{scheres2017plant} use the idea of a ``plant perceptron" to review recent findings about how plants integrate environmental signals with developmental processes to produce phenotypic variation, drawing an analogy between genetic and molecular signaling in plants and the signal integration of a perceptron in machine learning.

Other work has developed explicit mathematical models of plant behavior and growth. \citet{meroz2021plant} model plants as input-output systems that generate tropisms and reviewed models of plant shape formation, temporal integration of signals, and responses to neighboring plants. \citet{del2023perspectives} develop a perceptron-like model in which several stimulus-dependent behaviors contribute to plant growth and use this model to build bio-inspired control architectures. \citet{yang2009self} and \citet{feller2015mathematical} develop models of root-shoot growth and resource allocation. The former examines how shoot modules compete for root-derived resources, while the latter modeled balances root-shoot growth under varying environmental conditions. 

Some work has used plant phenomena as inspiration for solving optimization problems. Metaheuristic optimization is a generic framework for solving complex optimization problems using trial and error. \cite{akyol2017plant} extensively review 13 examples of existing metaheuristic optimization methods inspired by plant behavior.

Recent research has increasingly emphasized the distributed nature of plant behavior. \citet{bassel2018information} propose viewing plant organs as distributed computational systems in which individual cells process environmental information and communicate through molecular signals. \citet{duran2019plant} extend this perspective by considering how connected cellular units collectively control plant growth and development, while \citet{oborny2019plant} models plants as a network of semi-autonomous modules that exchange information and resources. 

The literature described above provides several complementary perspectives on plants, ranging from predictive processing and active inference to distributed cellular networks. However, this work primarily seeks to \textit{explain, model, or replicate specific forms of plant behavior}.  Our goal is different: rather than constructing computational models to explain plant behavior, we use plant behavior as a source of inspiration to identify new classes of AI problems. The resulting AI PFs, though inspired by plants, might abstract away details that would be of interest to biologists. Our PFs are intended to capture broader classes of computational problems than those found solely in plants. The two novel PFs that we present differ substantially from existing AI PFs: the first (Section \ref{ref:pf1}) because of how perception and morphology are intertwined, and the second (Section \ref{ref:pf2}) because of how resource allocation and morphology are interdependent. Appendix \ref{ref:appendix} lists further plant behaviors that may inspire similar formulations.

\begin{figure}
    \begin{subfigure}[c]{0.54\linewidth}
        \includegraphics[width=1.0\linewidth]{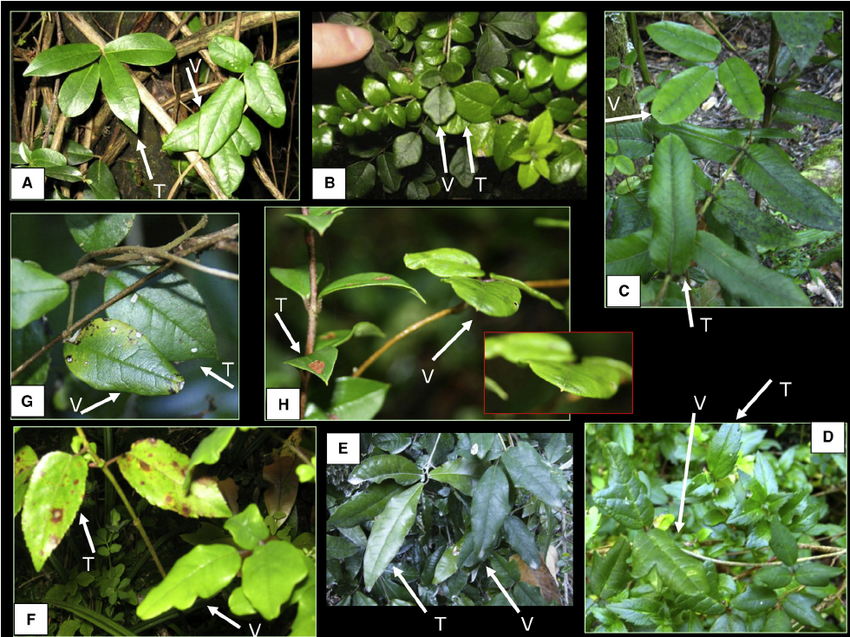} 
        \caption{}
        \label{fig:boquila}
    \end{subfigure}
    ~
    \begin{subfigure}[c]{0.46\linewidth}
        \includegraphics[width=1.0\linewidth]{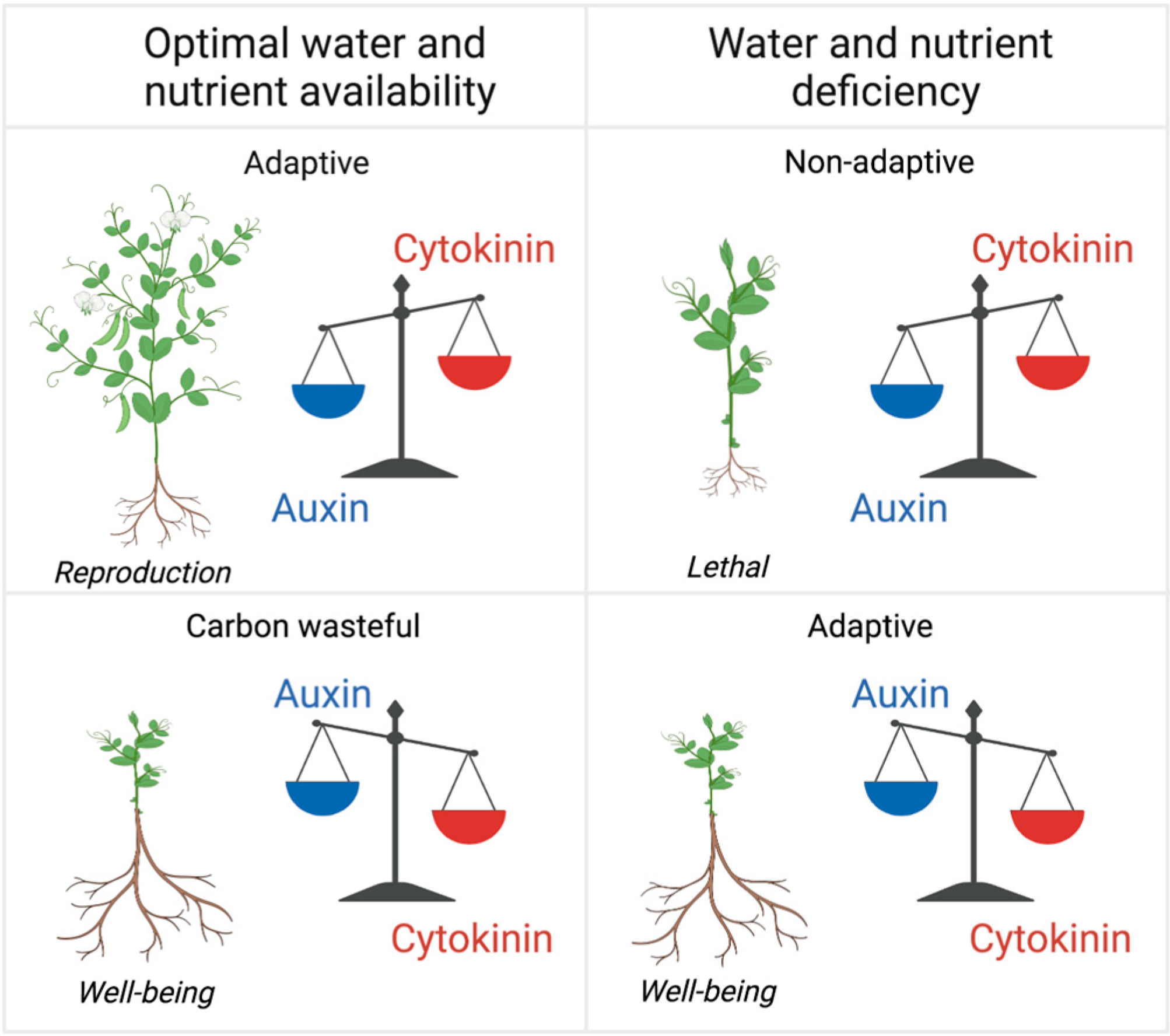}
        \caption{}
    \end{subfigure}
    \caption{Overview of the two case studies. (a) Examples of leaf mimicry in \textit{B. trifoliolata} \citep{gianoli2014leaf}. Host tree leaves are marked with \textbf{T}, while the \textit{Boquila} leaves are marked with \textbf{V}. (b) The ratio of auxin to cytokinin determines the extent to which roots and shoots grow in the plant \citep{kurepa2023friends}. Different environmental conditions result in different optimal ratios.}
    
\end{figure}

\section{Case Study 1: Leaf Mimicry in \textit{Boquila trifoliolata}} \label{ref:pf1}

\textit{B. trifoliolata}, also known as pil-pil or voquillo, is a climbing plant native to Chile and Argentina \citep{christenhusz2012738}. It has attracted recent attention for its ability to mimic the visual features of leaves of supporting trees \citep{gianoli2014leaf}. When leaves of other plant species are present, an individual plant can produce leaves that imitate those of multiple different host species simultaneously. By modifying leaf morphology, \textit{B. trifoliolata} can mimic more than a dozen host species, which reduces herbivory (Figure \ref{fig:boquila}).

How \textit{B. trifoliolata} achieves such sophisticated mimicry remains largely unclear. Proposed mechanisms include the detection of host trees' volatile organic compounds, horizontal gene transfer, and visual perception. Supporting the possibility of chemical interactions, bacterial communities in mimetic \textit{Boquila} leaves showed greater overlap with those of host tree leaves than do communities found in non-mimetic leaves \citep{gianoli2021endophytic}. More recently, \textit{Boquila} was discovered to be able to mimic plastic leaves, supporting the vision hypothesis \citep{white2022boquila}. Regardless of which mechanism ultimately explains mimicry in \textit{Boquila}, a separate computational problem remains: \textit{What kind of information do individual leaf cells detect and how do these cells communicate with each other to coordinate their growth so that the leaf as a whole becomes a successful mimic?} This communication and coordination problem is the basis of our first novel PF.

\subsection{Collective mimicry as a Problem Formulation}

\noskip \noindent \textbf{Structure:} There exists some target graph $\mathcal{T}$, and a similarity metric $M$ which is defined over $\mathcal{G}\times \mathcal{G}$, where $\mathcal{G}$ is the space of possible cell graphs. A mimic is an undirected graph of cells $G = (C, E)$. $C = \{c_{1:m}\}$, and $E$ is a $C \times C$ adjacency relation. Each cell $c$ has $(s, d, p)$: a state $s \in \{S\}$, a feature detector $d$ which perceives features of the target graph, and a position $p$.

\noskip \noindent \textbf{Cell Perception:}
Each cell may perceive some limited features of the target object, conditional on its position and timestep. A cell might or might not also receive a feedback signal, a recording of the global objective's completeness during the previous timestep. This feedback signal might be global or local to some cells.

\noskip \noindent \textbf{Cell Actions:} $A_c(t)$: At each timestep, a cell selects a set of actions $A_c(t)\subseteq\mathcal{A}$, where
\[
\mathcal{A}=\{ \texttt{create-cell}, \ \texttt{signal}\}.
\]
    The action \texttt{create-cell}: A cell $c_a$ may create a new cell $\hat{c}$ with edge ($\hat{c}$, $c_a$) $\in G$. The new cell has some position, state, and adjacencies with other cells. 
    Each cell may also \texttt{signal} to its neighbors and its future self the information it has accumulated about the visual features of the target object so other cells can contribute to collective mimicry.

\noskip \noindent \textbf{Objective:} The solution to this problem is maximizing similarity between the mimic and target: 
\[
G^*=\arg\max_{G} M(G,\mathcal{T})
\]

\subsection{Key Problem Characteristics}

In this section, we identify novel key problem characteristics (KPCs) of the mimicry PF. KPCs are often exposed at the level of PFs. For example, RL has its own set of KPCs, including exploration versus exploitation 
and temporal credit assignment. 
The mimicry problem features several KPCs which are common in other AI PFs, such as communication and information aggregation: To globally succeed at the mimicry task, cells must convey what they observe to other cells, and a meaningful representation of the host tree's leaf must be coalesced from many such observations. This representation could exist within individual cells or among a collective of cells.

\paragraph{Substrate Plasticity:}

New cells in this formulation become new feature detectors and signal communicators. Decisions about how to grow irreversibly alter the information available for all future decision-making. This means each new cell's placement must optimally satisfy several competing desiderata, including gathering information and signaling, in addition to the global goal of present or eventual mimicry. In \textit{Boquila}, the structure of a leaf likely places significant constraints on image processing and decision-making capabilities. Problems of this formulation cannot be approached with distinct solutions for perception and morphology.

\paragraph{Relative localization:}

Each individual cell behaves independently. To mimic some target, a cell in some location of the mimic must correspond with a location in the host. Individual cells must know their position in the leaf and how that appears in a corresponding host tree leaf, in order to decide how to grow. Even if the cell's state includes its location within the leaf, inferring the corresponding host-leaf location requires coordination with other cells regarding information about the target's layout. Even if feedback of the reward signal is incorporated in perceptions, identifying cells' individual effects on the goal could remain non-trivial.

\subsection{A Computational Collective Mimicry Task} \label{ref:mimic}

In this section, we operationalize the mimicry PF as a computational task in which population of cells must grow and coordinate their local decisions to mimic the structure of an unseen target.

\noskip \noindent \textbf{Task Specification}
A single \textit{cell} is instantiated. Cells have no absolute position or size, but exist on a grid relative to one another. Cells have a hidden state, which is a binary vector of length $n = 8$. A \textit{target image} is a previously unseen image selected from the MNIST dataset of 28$\times$28 grayscale handwritten digits \citep{lecun1998gradient}.

The objective is for the set of cells to grow such that their shape imitates the shape of a handwritten digit. Success is determined by comparing the target image (cropped to a zero-padded bounding box) to the shape of the cells at arbitrary resolution: both are treated as functions $T$ and $G$ in unit squares. At some comparison resolution, we sample the centers of pixels of either function. The error to be minimized, our similarity metric $M(G,T)$, is the L2 distance of these samples.

\begin{figure}[t!]
    \centering
    \includegraphics[width=0.98\linewidth]{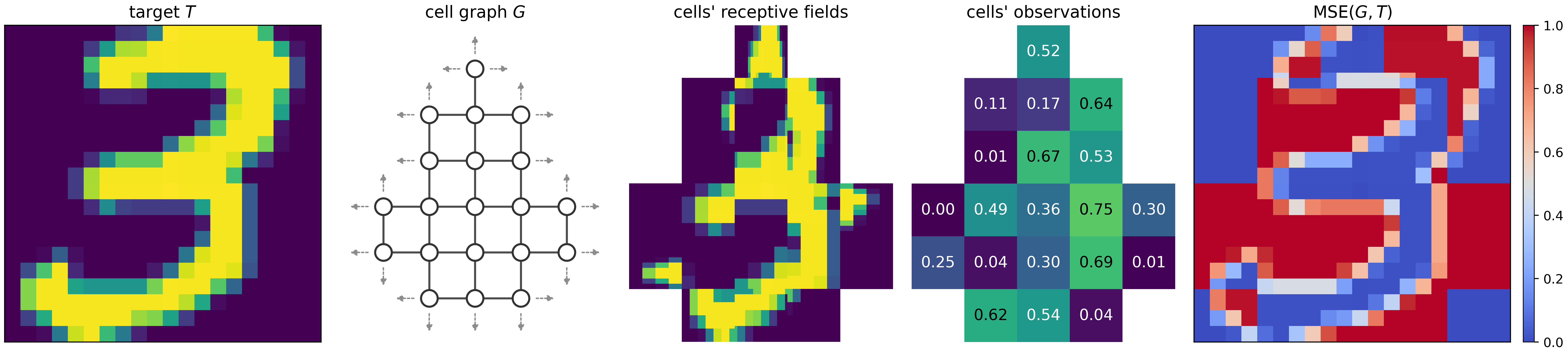}
    \caption{Instantiation of the collective mimicry task described in Section \ref{ref:mimic}. The left image is a target to be imitated, sampled from the MNIST dataset \citep{lecun1998gradient}. The second image shows a cell graph $G$ that might have been grown in an attempt to solve this task instance. This graph is not shaped like the digit depicted in $T$, so it is not a particularly good solution in this case. Edges between cells indicate the neighbors whose hidden states each cell may observe, while dotted arrows show where new cells may be grown. The central image shows the receptive fields of the cells; these fields depend on each cell's position relative to contiguous cells in its row and column. The fourth image shows the mean value of each field; these are the cells' observations of the target. Ultimately, the goal of the cells is to grow into the shape of the target image; the rightmost image shows the error in the appearance of the grown cell structure compared with the target $T$. }
    \label{fig:mimic_game}
\end{figure}

\noskip \noindent \textbf{Observations} Each cell is presented with an observation of the target image, and observations of its own plus 4 neighboring hidden states. The target image observation is a scalar sampling of the target image whose value depends on the cell's neighbors. Along either axis, contiguous groups of cells generate samplings of the target image. For example, a single cell on its own would receive only a mean valued observation of the entire target image. A row of four contiguous cells would generate a blurry (via linear interpolation) image with resolution four of the target 28-pixel image; each cell of these four would receive a quarter of the full image. Each cell also observes its own plus all four neighboring cells' hidden states of length $n$, resulting in an observation of size $1 + 5n$; at $n=8$ this size is 41. 

\noskip \noindent \textbf{Actions}
Each cell acts independently, simultaneously. Over 100 steps (flood-filling an MNIST image takes at least 14 steps), cells may: update their own hidden state, and/or create a new cell in any unoccupied adjacent tiles. These actions may be selected with $4 + N$ binary output units, making the action space size $16 \times 2^{n}$; at $n=8$ this size is 4096 for each cell.

\paragraph{Why this task is challenging?} This task heavily emphasizes cells' ability to communicate given extremely impoverished imaging capabilities. A single cell has almost no useful information about the image to be imitated, and cells in groups do not directly access useful information about how they should grow. 

Solving this task also requires overcoming the novel limitations of substrate plasticity and irreversible commitment. Actions create new sensors, and sensors' receptive fields are determined by previous growth. Because cells in this task do not change after creation, every action shrinks the action space of neighboring cells and permanently constrains the eventual output.

\paragraph{What is Missing?}
This test does not incorporate feedback, which could be a useful component for \textit{Boquila}'s solution to the problem of mimicry. Given such a signal, inter-cellular communication could be otherwise unnecessary, since cells could perturb their states and leverage memory to solve the problem via evolution or a localized gradient descent. The target image in this setup also has a fixed location relative to the sensors, so image segmentation or object tracking are not needed to solve the problem of relative localization. All cells in this task broadcast their state to all their neighbors, while in plants communication channels are likely more selective. Additionally, real world constraints might differentiate communication from memory; our hidden state vectors simultaneously encompass both communication and memory. Placing further constraint on these hidden states, e.g. by making a subset of the state visible to neighbors, could reveal the importance of disentangling communication from memory. 

\paragraph{Hypothetical Solutions}
A hardcoded solution to this task might involve explicit communication channels describing cells' observations and whereabouts using lengthy hidden state vectors. Given unlimited hidden states and time, each cell in a large enough mimic could accumulate enough information to represent the full target image, and could localize its own relative position in the target image. The task as written, however, places severe limitations on hidden state size and the amount of time for message passing. Solving it requires more efficiently encoded communication and representations. A rough analogy can be found in the stigmergy of ants, who alter the environment through their interactions and then make decisions based on locally sensed pheromones \citep{theraulaz1999brief}. These local cues then give rise to macro-level structures like bridges.

A solution could leverage RL algorithms designed for multi-agent partially-observable Markov decision processes carried out over many timesteps. It is unclear, however, that off-the-shelf setups are able to overcome the extreme limitations of observability, the complexity of communication required, and the largely unexplored challenges involving substrate plasticity and relative localization. 

This problem closely resembles the work presented in \cite{mordvintsev2020growing}. In this work, cells coordinate to grow and robustly repair given images. However, in our collective mimicry task, perception is elevated to a key challenge; the goal is mimicry of any target image rather than a fixed image, and cells must coordinate to perceive any useful information about the image.

\section{Case Study 2: Collaborative Growth of Roots and Shoots} \label{ref:pf2}

Most plants grow in two distinct environments simultaneously: the soil and the air. This growth is organized into two interconnected systems, the roots and shoots, which develop from their own populations of apical meristems (stem cells). The two systems specialize in acquiring complementary resources: roots acquire water and mineral nutrients from the soil, while shoots capture light and atmospheric CO$_{2}$ for photosynthesis. Maintaining an optimal balance between root and shoot growth is critical for plant survival and reproduction, but the appropriate balance depends on environmental conditions and resource availability. This balance is regulated by a complex set of hormonal signals, in particular, auxin and cytokinins that promote root and shoot growth respectively. Mathematical models have examined how local physiological processes and hormonal signaling can coordinate growth and resource allocation \citep{yang2009self,feller2015mathematical}.

Roots and shoots solve distinct resource acquisition problems through local sensing and adaptation. Roots explore the spatiotemporally heterogenous soil environment to locate nutrients: auxin redistribution alters growth on different sides of the root causing it to bend towards water, membrane transporters detect nutrients like nitrates and phosphates, hormone signaling regulates the growth of lateral roots, and specialized statocyte cells detect gravity. 

Shoots, in turn, respond to the spatial and temporal structure of the atmospheric environment, adjusting growth in response to light intensity and spectral composition, neighboring vegetation, temperature, and humidity. Yet, these two organ systems are tightly coupled. Carbohydrates generated in shoots as a result of photosynthesis support root growth, while water and mineral nutrients acquired by roots support photosynthesis in the shoots. Thus, whole-plant growth emerges from interactions between two spatially separated populations of specialized components that must continuously acquire, allocate, and exchange resources.  

\subsection{Problem Formulation}

\noskip \noindent \textbf{Environment} $E$: The environment encodes the spatiotemporal distribution of resources, $\rho(\mathbf{x}, t)$, in the environment in which the AI (plant) operates.

\noskip \noindent \textbf{Cell complex} $C_t$: The cell complex at time $t$ is the set of all living cells, $C_t=\{c_1,\ldots,c_{N_t}\}$, where $N_t=|C_t|$ is the number of cells. Each cell is characterized by its spatial position, neighboring cells, cell type, resource buffer, and additional internal state variables. 
    
\noskip \noindent \textbf{Cell} $c$: Each cell has a fixed spatial position $\mathbf{x}=(x,y,z)$. If the cell $c$ is created at time $t_c$, it can sense the local environment, $\rho(\mathbf{x}, t)$, for all $t \geq t_c$. At each time step $t$, the cell receives the incoming signal set 
\[
S_c(t-1)=\{s_{j\rightarrow c}(t-1)\},
\]
where $s_{j\rightarrow c}(t-1)$ denotes the signal sent from cell $j$ to cell $c$ at the previous time step. These signals may include both information and resources transferred from other cells.

\noskip \noindent \textbf{Cell Maintenance Cost}: Each living cell incurs a maintenance cost that must be paid from its resources. A cell unable to meet the maintenance cost dies and is removed from the cell complex.

\noskip \noindent \textbf{Cell resources} $\mathbf{r}_c(t)$: Each cell maintains a resource buffer $\mathbf{r}_c(t)\in\mathbb{R}_{\ge0}^m$, where $m$ is the number of resource types. The resource buffer is updated over time through acquisition from the environment, transfer from other cells, and expenditure on cell's actions and maintenance.

\noskip \noindent \textbf{Cell actions} $A_c(t)$: At each time step, a cell selects a set of actions $A_c(t)\subseteq\mathcal{A}$, where
\[
\mathcal{A}=\{\texttt{signal},\ \texttt{create-cell}\}.
\]
The action \texttt{signal} transmits one or more signals $s_{c\rightarrow j}(t)$ to other cells and may decrease $\mathbf{r}_c(t)$ if resources are transferred. The action \texttt{create-cell} creates a new cell $c'$ at position $\mathbf{x}'$, provided that $\mathbf{r}_c(t)$ contains sufficient resources to satisfy the resource requirements for cell creation. If $A_c(t)=\emptyset$, the cell takes no action during time step $t$.

\noskip \noindent \textbf{Utility metric} $H(U_t,E)$: A scalar-valued function $H:\mathcal{U}\times\mathcal{E}\rightarrow\mathbb{R}$ that evaluates the quality of the current cell complex in environment $E$. The metric may depend on properties such as the number and distribution of cell types, spatial organization, resource acquisition and allocation, and the overall size of the cell complex.

\subsection{Key Problem Characteristics}

This PF shares many features of the mimicry PF, including a growing graph of interacting components, decentralized communication and coordination among local decisions. However, this PF introduces additional challenges arising from resource dependencies and heterogenous cell roles.

\paragraph{Resource Allocation:} Unlike the mimicry problem, this PF requires the allocation of multiple resources across a growing structure. Resources have different spatiotemporal distributions, and individual cells can acquire only a subset of available resources. Cells must therefore decide when to acquire, store, transfer and consume resources. The maintenance costs of cells introduces an additional tradeoff between maintaining the existing structure and investing resources for further growth. Cell death thus provides a mechanism through which inefficient cells can be pruned, making population maintenance itself part of the decision problem.

\paragraph{Heterogeneous Roles:} The cell complex contains heterogeneous cell types with distinct capabilities, resource-access constraints, and growth requirements. 
Consequently, no single cell can independently acquire all resources required for continued growth. Successful growth requires distinct local policies for different cell types and coordination between them so locally acquired resources are distributed to support the maintenance and growth of the entire cell complex.

\subsection{A Computational Root-Shoot Collaboration Task} \label{ref:rootsnshoots}
We instantiate the root-shoot collaboration PF as a benchmark in which a growing population of heterogenous cells must acquire, store, and exchange resources to maximize the growth of the overall cell complex. The benchmark does not attempt to reproduce the biochemical mechanisms of real plants, but instead isolates the computational structure identified by the PF.

\begin{figure}[t!]
    \centering
    \includegraphics[width=1.0\linewidth]{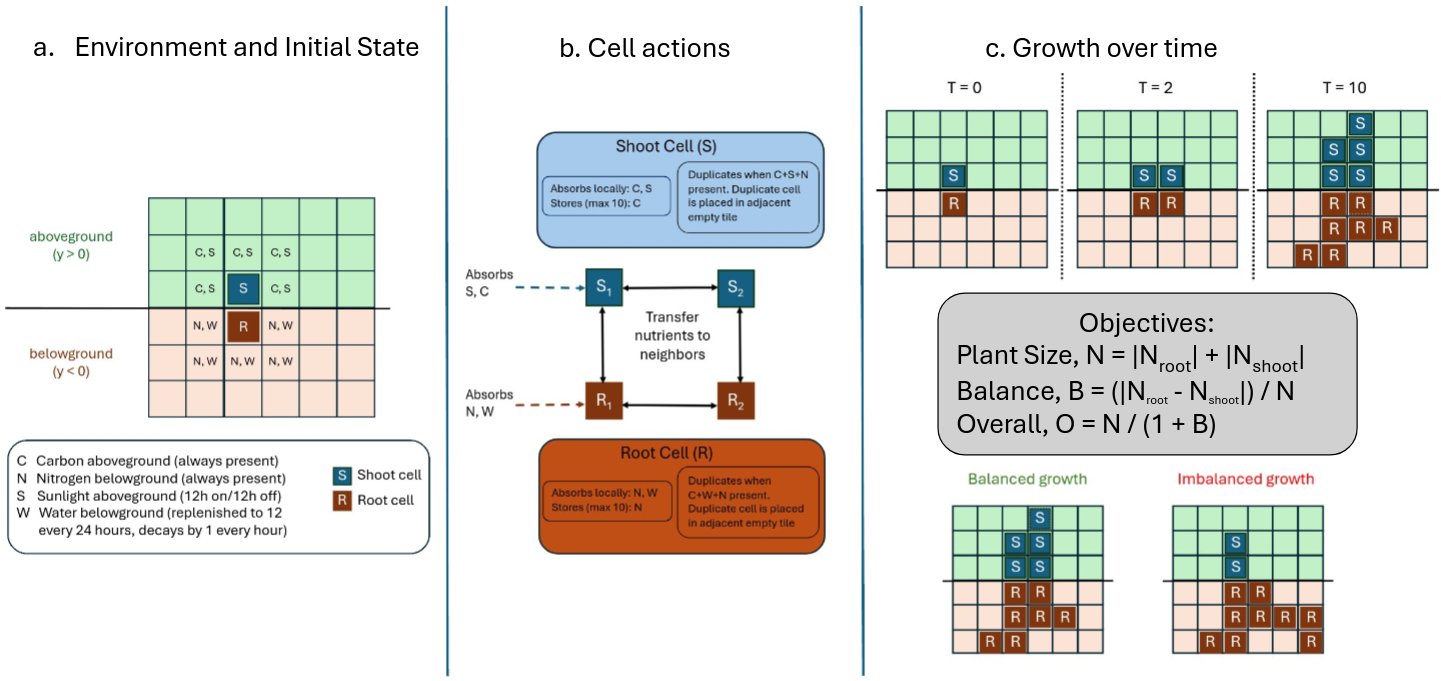}
    \caption{\textbf{Overview of the root-shoot collaboration task.} (a) The environment is divided into aboveground and belowground regions with different nutrient distributions. A shoot cell $S$ can acquire Carbon and Sunlight, while a root cell $R$ can acquire Nitrogen and Water. (b) Cells sense and store locally available nutrients, transfer nutrients to neighboring cells, and duplicate when growth requirements are met. (c) These local actions produce coordinated growth of the root and shoot over time, with growth being measured by the plant size $N$ and balance $B$.}
    \label{fig:root_shoot_task}
\end{figure}

\noskip \noindent \textbf{Initial conditions} 
    The cell complex initially consists of one root cell and one shoot cell. Cells can create additional cells when their growth requirements are satisfied. Resources transferred between root and shoot tissues are distributed among cells in the receiving tissue. 

\noskip \noindent \textbf{Environment and nutrients} 
    The environment is represented as a grid of tiles, with different nutrients available at different location. We consider four nutrient types: \textit{Carbon}, \textit{Sunlight}, \textit{Nitrogen}, and \textit{Water}. \textit{Carbon} is continuously available at a constant density aboveground ($y > 0$), while \textit{Nitrogen} is continuously available at a constant density below-ground ($y < 0$). \textit{Sunlight} follows a 24-hour cycle, remaining available for 12 hours and unavailable for the next 12 hours. \textit{Water} is replenished to a value of 12 every 24 hours in each below-ground tile and decays by 1 each hour. 
    
    Each cell can absorb at most one unit of nutrient per timestep, removing the absorbed unit from its tile. Shoot cells can absorb \textit{Carbon} and \textit{Sunlight} and may store up to 10 \textit{Carbon}, while Root cells can absorb \textit{Nitrogen} and \textit{Water} and may store up to 10 \textit{Nitrogen}.

\noskip \noindent \textbf{Actions} 
    At each timestep, a cell may transfer one unit of each stored nutrient to any of its neighboring cells. Cells may also create a new cell by filling an empty neighboring tile. A root cell can create a new cell only when \textit{Carbon}, \textit{Water}, and \textit{Nitrogen} are simultaneously available, while a shoot cell can create a new cell only when \textit{Sunlight}, \textit{Nitrogen}, and \textit{Carbon} are simultaneously available. Thus, each cell type can locally acquire two of the resources required for growth but must obtain the third through the broader cell complex.

\noskip \noindent \textbf{Objective} 
    We evaluate the current state of the cell complex based on the size and balance between the root and shoot cells in the complex. Total plant size, $N$ is defined as $N = N_{\text{root}} + N_{\text{shoot}}$, where $N_{\text{root}}$ and $N_{\text{shoot}}$ denote the numbers of living root and shoot cells. We measure growth imbalance $B$ as
    $B = |N_{\text{root}} - N_{\text{shoot}}| / |N_{\text{root}} + N_{\text{shoot}}|$ and combine these quantities into the objective $O = N / (1 + B)$. The benchmark asks agents to design or learn separate policies for root and shoot cells that use local resource sensing, storage, transfer and communication to maximize the holistic growth of the cell complex.

\paragraph{Why is this a challenging task?}
Root and shoot cells have asymmetric resource requirements and access to different subsets of environmental resources. Additionally, growth in one type of cell depends on resources transferred from the other, creating dependencies across the two populations. As a result, locally optimal decisions can be globally harmful: for example, allocating resources to create new root cells may improve future nutrient acquisition, but reduce resources available for shoot growth. Cells must therefore learn when to store, transfer, or consume resources for growth while coordinating with other cells as the structure of the cell complex evolves over time.

\paragraph{What is missing in this task?}
The task omits several aspects of biological growth. It does not require spatial foraging since the cell complex does not need to decide where to grow in response to spatially heterogeneous resource availability. Additionally, the environment imposes very few structural restrictions on where new cells can grow: cells can grow without modeling physical support, overcrowding, or other morphological constraints. These omissions simplify the problem, but incorporating foraging and  morphological limitations can make progressively harder tests of decentralized growth.

\paragraph{Hypothetical Solutions}
One possible solution is for each cell to maintain an internal model of the resource state and requirements of other parts of the cell complex. Such a model could allow cells to infer whether another cell is likely to experience a resource deficit and handle the tradeoff between resource transfer and local growth. Although the PF and the task are inspired by plant behavior, the solution algorithms used to solve this task need not resemble mechanisms used by plants. 

Additionally, this is only one possible solution strategy. The PF does not prescribe adherence to any solution strategy, be it centralized planning, a specific communication protocol, or even an explicit representation of other cells. Alternatively, cells could employ simple local policies that transfer resources based only on local observations and incoming signals, with global growth emerging from their repeated interactions. Comparing such solution strategies would allow us to study how different forms of coordination can solve the same underlying collaboration problem.

\section{Related Work: Similar PFs in AI}

\noskip \noindent \textbf{Reinforcement Learning}
The Markov Decision Process (MDP) formalism from RL has been extended to study collaborative decision-making among decentralized groups of agents. \citet{bernstein2002complexity} introduce extensions of both MDPs and Partially Observable MDPs to include decentralized control and demonstrated differences in computational complexity compared to centralized control. Subsequent work has incorporated additional structure into decentralized decision-making, including networked interactions \citep{nair2005networked}, locality exploitation in POMDPs \citep{oliehoek2008exploiting}, and explicit communication among agents \citep{tasaki2010introducing}. \citet{amato2013decentralized} provides a detailed survey of different POMDP variants and their computational complexities.

\noskip \noindent \textbf{Cellular Automata}
Another relevant line of research comes from Cellular Automata (CA). Recent work in CA has explored how local rules can generate complex biological behavior. Biomaker CA \citep{randazzo2023biomaker} uses CA to model plants that grow and reproduce. Neural CA \citep{mordvintsev2020growing} learns local rules that can grow and regenerate complex target images.

\noskip \noindent \textbf{Autoencoders}
Autoencoders are machine learning models which compress some input and then reconstruct the original from its compressed encoding \citep{hinton2006reducing}. They are widely used for representation learning and dimensionality reduction. Our mimicry PF shares the same goal: a target image is perceived, internally represented, and then reconstructed by the model. Autoencoders have a fixed architecture, whereas in our PF the architecture itself is the output. 

\noskip \noindent \textbf{Superresolution}
Super-resolution describes the reconstruction of a high-resolution image from one or more low-resolution observations. In multi-frame super-resolution, a high-resolution image is produced by leveraging multiple low-resolution observations that differ in their spatial sampling of the image \citep{irani1991improving}. This problem shares a core mystery of \textit{Boquila} behavior: how does it know what the host leaf looks like, especially when leaves move and differ in angular displacement? Multi-frame super-resolution assumes all frames are collected before being aligned, while our PF intentionally excludes the guarantee of simultaneous access to the full set of frames.

\section{Discussion} \label{ref:discussion}

We have presented two distinct PFs inspired by the complex behavior that plants demonstrate. While both PFs have overlapping KPCs, each introduces novel challenges. The design of the PFs constrains how they might map onto different sets of problems and how they might be solved. While we opted to represent both problems as a search for graph structures, we might have designed systems that are continuous, or that are necessarily embedded in a physical environment.  

In developing these case studies, we found that, to our knowledge, there exists no established method for evaluating the quality of a PF. This absence is notable considering the importance of PFs to AI and the rigor with which specific tasks and solution algorithms are studied. 
Ultimately, PFs are validated retrospectively through adoption by other scientists and their usefulness for solving important problems. However, several properties may contribute to initial design of a PF: the ability to naturally express diverse real-world problems, capturing key structure not naturally present in other PFs, and sufficient structure to enable effective solution algorithms.

Our first PF, collective mimicry, involves leveraging information gathered by existing components to determine where new information-gathering components should be grown. Several real-world applications have a similar structure: in software engineering, import conflicts in code libraries could be resolved by generating code bases imitating the necessary functionality; in urban design, designs could be iteratively generated to approximate high-level concepts while adhering to spatial and resource constraints. 
Our second PF, root-shoot collaboration, involves coordination and resource allocation between heterogeneous components that acquire different resources locally. Real world applications include designing robust communications networks like internet service provider infrastructure, or feedback-regulated transition to solar and other renewable energies.

We are particularly curious about the extent to which the KPCs that we have identified are solved by existing AI techniques. The development of RL, for example, drew not only from animal learning but also from existing research in dynamic programming and optimal control, with these ideas eventually contributing to methods like Q-learning. Similarly, solutions for our PFs may benefit from existing work in optimization, graph algorithms, and adaptive control, or may require novel combinations of these techniques. 
Our PFs were selected around the KPCs that we found particularly striking in the motivating biological phenomena. Collective mimicry makes distributed information gathering and growth explicit, while root-shoot collaboration makes coordinated resource acquisition, resource allocation and heterogeneous roles explicit. Our PFs and task designs place these KPC at the forefront, making them first-order problems. 

Research on plant behavior shares some similarities with pre-1950s research on human behavior. There are rich theories about plants at the behavioral level, e.g., describing how they respond to stimuli in different environments. There are also rich studies at the physiological level, describing genetic, molecular, and cellular processes.  In contrast, just as was the case in pre-1950s human behavioral research, there are dramatically fewer theories that explain plant behavior in terms of underlying representations and algorithm-like processes. Plant-inspired AI may therefore offer a way to investigate not only what plants do, but what kinds of computational architectures could make their behaviors possible.

\citet{levin2026machines} argue for viewing cognitive capabilities as a continuum across biological scales, from cellular processes to organism behavior. These different organizational levels can account for different cognitive capacities and can be useful for studying intelligent behavior at different scales. We view plant-inspired AI in this broader sense. While our two case studies both relied on decentralized, cell-based computations as the underlying abstraction, considering plant intelligence at different organization levels (e.g., cellular, organ, whole-plant, and plant-environment scales) can inspire other types of PFs.

\vspace{-0.15in}

\begin{acknowledgements} 
\noindent
We would like to thank the consultants of the PLACS/PLAI meeting for their contributions to this work, including Eric Brenner, David Crandall, Ole Molvig, Katja Tielb\"orger, Elizabeth Van Volkenburgh, and Emanuela Sani. This work was supported by a grant from the John Templeton Foundation to M.K. and A.B.R.
\end{acknowledgements}

\vspace{-0.15in}

{\parindent -10pt\leftskip 10pt\noindent
\bibliographystyle{cogsysapa}
\bibliography{format}

}
\newpage
\section{Appendix}\label{ref:appendix}

\subsection{Further examples of Plant Problem Solving}

Further future work will explore more of the breadth of plant phenomena which might inspire new PFs. In this appendix, we list several examples of plant behavior which may inspire new tasks or PFs.

\paragraph{Common to most or virtually all plants}

Plants perform arithmetic at night by dividing their starch reserves by the remaining hours of darkness to pace their metabolic rate so they run out of energy at sunrise, and recalculate this rate if darkness falls early \citep{scialdone2013arabidopsis}. 

To manage limited carbon stores, plants flexibly employ a modular defense strategy---sacrificing doomed, heavily shaded leaves by suppressing their immune responses and redirecting resources towards vertical growth and immune defenses in valuable, sunlit leaves\citep{ballare2014light}.

When the spectral quality of upward-reflected light from leaves indicates heavy lower-canopy shading (a low red to far-red ratio), upper shoots orient their leaves more vertically, dynamically engineering the canopy architecture to allow deeper light penetration \citep{zhang2021light}.

\paragraph{Common to broad groups of species}

Fast-growing herbaceous plants prefer to grow roots in predictable nutrient conditions when resources are adequate, but gambling on highly variable, unpredictable soil patches when starved of nutrients \citep{dener2016pea}.

Tropical climbing vines survive the jungle understory by initially growing toward the darkest shadows to locate the largest host tree (skototropism), and then immediately reversing their strategy upon contact to grow upwards toward the light \citep{strong1975host}.

Plants capable of dimorphic cleistogamy, or mixed-mating systems, monitor their environment to produce large flowers for insect cross-pollination when conditions are favorable, but switch to producing small, closed, self-pollinating flowers in less favorable conditions \citep{culley2007cleistogamous}.

\paragraph{Single species or small subset of species}

To avoid starvation within days of germination, rootless parasitic Dodder vines evaluate the airborne chemical volatiles of nearby plants to make a cost-benefit choice between the nutrition quality and reachability of potential hosts \citep{runyon2006volatile}.

A Venus flytrap performs complex arithmetic by requiring two distinct hair touches by an insect within thirty seconds to snap its trap shut, waiting for five touches by an insect to produce  digestive enzymes, and counting subsequent insect touches to proportionally scale the production of these enzymes to the active struggle level (number of touches) of the trapped prey. \citep{bohm2016venus}.


\end{document}